# A Counterexample to Theorems of Cox and Fine

**Joseph Y. Halpern**                                        HALPERN@CS.CORNELL.EDU
*Cornell University, Computer Science Department*
*Ithaca, NY 14853*
*http://www.cs.cornell.edu/home/halpern*

## Abstract

Cox's well-known theorem justifying the use of probability is shown not to hold in finite domains. The counterexample also suggests that Cox's assumptions are insufficient to prove the result even in infinite domains. The same counterexample is used to disprove a result of Fine on comparative conditional probability.

## 1. Introduction

One of the best-known and seemingly most compelling justifications of the use of probability is given by Cox (1946). Suppose we have a function Bel that associates a real number with each pair $(U, V)$ of subsets of a domain $W$ such that $U \neq \emptyset$. We write $\text{Bel}(V|U)$ rather than $\text{Bel}(U, V)$, since we think of $\text{Bel}(V|U)$ as the credibility or likelihood of $V$ given $U$.[1] Cox further assumes that $\text{Bel}(\overline{V}|U)$ is a function of $\text{Bel}(V|U)$ (where $\overline{V}$ denotes the complement of $V$ in $W$), that is, there is a function $S$ such that

**A1.** $\text{Bel}(\overline{V}|U) = S(\text{Bel}(V|U))$ if $U \neq \emptyset$,

and that $\text{Bel}(V \cap V'|U)$ is a function of $\text{Bel}(V'|V \cap U)$ and $\text{Bel}(V|U)$, that is, there is a function $F$ such that

**A2.** $\text{Bel}(V \cap V'|U) = F(\text{Bel}(V'|V \cap U), \text{Bel}(V|U))$ if $V \cap U \neq \emptyset$.

Notice that if Bel is a probability function, then we can take $S(x) = 1 - x$ and $F(x, y) = xy$. Cox makes much weaker assumptions: he assumes that $F$ is twice differentiable, with a continuous second derivative, and that $S$ is twice differentiable. Under these assumptions, he shows that Bel is isomorphic to a probability distribution in the sense that there is a continuous one-to-one onto function $g : I\!R \to I\!R$ such that $g \circ \text{Bel}$ is a probability distribution on $W$, and

$$g(\text{Bel}(V|U)) \times g(\text{Bel}(U)) = g(\text{Bel}(V \cap U)) \text{ if } U \neq \emptyset, \qquad (1)$$

where $\text{Bel}(U)$ is an abbreviation for $\text{Bel}(U|W)$.

Not surprisingly, Cox's result has attracted a great deal of interest, particularly in the maximum entropy community and, more recently, in the AI community. For example

---

1. Cox writes $V|U$ rather than $\text{Bel}(V|U)$, and takes $U$ and $V$ to be propositions in some language rather than events, i.e., subsets of a given set. This difference is minor—there are well-known mappings from propositions to events, and vice versa. I use events here since they are more standard in the probability literature.

- Cheeseman (1988) has called it the "strongest argument for use of standard (Bayesian) probability theory". Similar sentiments are expressed by Jaynes (1978, p. 24); indeed, Cox's Theorem is one of the cornerstones of Jaynes' recent book (1996).

- Horvitz, Heckerman, and Langlotz (1986) used it as a basis for comparison of probability and other nonprobabilistic approaches to reasoning about uncertainty.

- Heckerman (1988) used it as a basis for providing an axiomatization for belief update.

The main contribution of this paper is to show (by means of an explicit counterexample) that Cox's result does not hold in finite domains, even under strong assumptions on $S$ and $F$ (stronger than those made by Cox and those made in all papers proving variants of Cox's results). Since finite domains are arguably those of most interest in AI applications, this suggests that arguments for using probability based on Cox's result—and other justifications similar in spirit—must be taken with a grain of salt, and their proofs carefully reviewed. Moreover, the counterexample suggests that Cox's assumptions are insufficient to prove the result even in infinite domains.

It is known that some assumptions regarding $F$ and $S$ must be made to prove Cox's result. Dubois and Prade (1990) give an example of a function Bel, defined on a finite domain, that is not isomorphic to a probability distribution. For this choice of Bel, we can take $F(x, y) = \min(x, y)$ and $S(x) = 1 - x$. Since min is not twice differentiable, Cox's assumptions block the Dubois-Prade example.

Other authors have made different assumptions. Aczél (1966, Section 7 (Theorem 1)) does not make any assumptions about $F$, but he does make two other assumptions, each of which block the Dubois-Prade example. The first is that the Bel$(V|U)$ takes on every value in some range $[e, E]$, with $e < E$. In the Dubois-Prade example, the domain is finite, so this certainly cannot hold. The second is that if $V$ and $V'$ are disjoint, then there is a continuous function $G : \mathbb{R}^2 \to \mathbb{R}$, strictly increasing in each argument, such that

**A3.** Bel$(V \cup V'|U) = G(\text{Bel}(V|U), \text{Bel}(V'|U))$.

With these assumptions, he gives a proof much in the spirit of that of Cox to show that Bel is essentially a probability distribution. Dubois and Prade point out that, in their example, there is no function $G$ satisfying A3 (even if we drop the requirement that $G$ be continuous and strictly increasing in each argument).[2]

Reichenbach (1949) earlier proved a result similar to Aczél's, under somewhat stronger assumptions. In particular, he assumed A3, with $G$ being $+$.

Other variants of Cox's result have also been considered in the literature. For example, Heckerman (1988) and Horvitz, Heckerman, and Langlotz (1986) assume that $F$ is continuous and strictly increasing in each argument and $S$ is continuous and strictly decreasing. Since min is not strictly continuous in each argument, it fails this restriction too.[3] Aleliunas (1988) gives yet another collection of assumptions and claims that they suffice to guarantee that Bel is essentially a probability distribution.

---

2. In fact, Aczél allows there to be a different function $G_U$ for each set $U$ on the right-hand side of the conditional. However, the Dubois-Prade example does not even satisfy this weaker condition.

3. Actually, the restriction that $F$ be strictly increasing in each argument is a little too strong. If $e = \text{Bel}(\emptyset)$, then it can be shown that $F(e, x) = F(x, e) = e$ for all $x$, so that $F$ is not strictly increasing if one of its arguments is $e$.

The first to observe potential problems with Cox's result is Paris (1994). As he puts it, "Cox's proof is not, perhaps, as rigorous as some pedants might prefer and when an attempt is made to fill in all the details some of the attractiveness of the original is lost." Paris provides a rigorous proof of the result, assuming that the range of Bel is contained in $[0, 1]$ and using assumptions similar to those of Horvitz, Heckerman, and Langlotz. In particular, he assumes that $F$ is continuous and strictly increasing in $(0, 1]^2$ and that $S$ is decreasing. However, he makes use of one additional assumption that, as he himself says, is not very appealing:

**A4.** For all $0 \leq \alpha, \beta, \gamma \leq 1$ and $\epsilon > 0$, there are sets $U_1 \supseteq U_2 \supseteq U_3 \supseteq U_4$ such that $U_3 \neq \emptyset$, and each of $|\text{Bel}(U_4|U_3) - \alpha|$, $|\text{Bel}(U_3|U_2) - \beta|$, and $|\text{Bel}(U_2|U_1) - \gamma|$ is less than $\epsilon$.

Notice that this assumption forces the range of Bel to be dense in $[0, 1]$. This means that, in particular, the domain $W$ on which Bel is defined cannot be finite.

Is this assumption really necessary? Paris suggests that Aczél needs something like it. (This issue is discussed in further detail below.) The counterexample of this paper gives further evidence. It shows that Cox's result fails in finite domains, even if we assume that the range of Bel is in $[0, 1]$, $S(x) = 1 - x$ (so that, in particular, $S$ is twice differentiable and monotonically decreasing), $G(x, y) = x + y$, and $F$ is infinitely differentiable and strictly increasing on $(0, 1]^2$. We can further assume that $F$ is commutative, $F(0, x) = F(x, 0) = 0$, and that $F(x, 1) = F(1, x) = x$. The example emphasizes the point that the applicability of Cox's result is far narrower than was previously believed. It remains an open question as to whether there is an appropriate strengthening of the assumptions that does give us Cox's result in finite settings. There is further discussion of this issue in Section 5.

In fact, the example shows even more. In the course of his proof, Cox claims to show that $F$ must be an associative function, that is, that $F(x, F(y, z)) = F(F(x, y), z)$. For the Bel of the counterexample, there can be no associative function $F$ satisfying A2. It is this observation that is the key to showing that there is no probability distribution isomorphic to Bel.

What is going on here? Actually, Cox's proof just shows that $F(x, F(y, z)) = F(F(x, y), z)$ only for those triples $(x, y, z)$ such that, for some sets $U_1$, $U_2$, $U_3$, and $U_4$, we have $x = \text{Bel}(U_4|U_3 \cap U_2 \cap U_1)$, $y = \text{Bel}(U_3|U_2 \cap U_1)$, and $z = \text{Bel}(U_2|U_1)$. If the set of such triples $(x, y, z)$ is dense in $[0, 1]^3$, then we conclude by continuity that $F$ is associative. The content of A4 is precisely that the set of such triples is dense in $[0, 1]^3$. Of course, if $W$ is finite, we cannot have density. As my counterexample shows, we do not in general have associativity in finite domains. Moreover, this lack of associativity can result in the failure of Cox's theorem.

A similar problem seems to exist in Aczél's proof (as already observed by Paris (1994)). While Aczél's proof does not involve showing that $F$ is associative, it does involve showing that $G$ is associative. Again, it is not hard to show that $G$ is associative for appropriate triples, just as is the case for $F$. But it seems that Aczél also needs an assumption that guarantees that the appropriate set of triples is dense, and it is not clear that his assumptions do in fact guarantee this.[4] As shown in Section 2, the problem also arises in Reichenbach's proof.

The counterexample to Cox's theorem, with slight modifications, can also be used to show that another well-known result in the literature is not completely correct. In his seminal book on probability and qualitative probability (1973), Fine considers a non-numeric notion of *comparative (conditional) probability*, which allows us to say "$U$ given $V$ is at least as probable as $U'$ given $V'$", denoted $U|V \succeq U'|V'$. Conditions on $\succeq$ are given that are claimed to force the existence of (among other things) a function Bel such that $U|V \succeq U'|V'$ iff $\text{Bel}(U|V) \geq \text{Bel}(U'|V')$ and an associative function $F$ satisfying A2. (This is Theorem 8 of Chapter II in (Fine, 1973).) However, the Bel defined in my counterexample to Cox's theorem can be used to give a counterexample to this result as well.

Interestingly, this is not the first time a similar error has been noted in the use of functional equations. Falmagne (1981) gives another example (in a case involving a utility model of choice behavior) and mentions that he knows "of at least two similar examples in the psychological literature".

The remainder of this paper is organized as follows. In the next section there is a more detailed discussion of the problem in Cox's proof. The counterexample to Cox's theorem is given in Section 3. The following section shows that it is also a counterexample to Fine's theorem. Section 5 concludes with some discussion, particularly of assumptions under which Cox's theorem might hold.

## 2. The Problem With Cox's Proof

To understand the problems with Cox's proof, I actually consider Reichenbach's proof, which is similar in spirit Cox's proof (it is actually even closer to Aczél's proof), but uses some additional assumptions, which makes it easier to explain in detail. Aczél, Cox, and Reichenbach all make critical use of functional equations in their proof, and they make the same (seemingly unjustified) leap at corresponding points in their proofs.

In the notation of this paper, Reichenbach (1949, pp. 65–67) assumes (1) that the range of $\text{Bel}(\cdot|\cdot)$ is a subset of $[0,1]$, (2) $\text{Bel}(V|U) = 1$ if $U \subseteq V$, (3) that if $V$ and $V'$ are disjoint, then $\text{Bel}(V \cup V'|U) = \text{Bel}(V|U) + \text{Bel}(V'|U)$ (thus, he assumes that A3 holds, with $G$ being $+$), and (4) that A2 holds with a function $F$ that is differentiable. (He remarks that the result holds even without assumption (4), although the proof is more complicated; Aczél in fact does not make an assumption like (4).)

Reichenbach's proof proceeds as follows: Replacing $V'$ in A2 by $V_1 \cup V_2$, where $V_1$ and $V_2$ are disjoint, we get that

$$\text{Bel}(V \cap (V_1 \cup V_2)|U) = F(\text{Bel}(V_1 \cup V_2|V \cap U), \text{Bel}(V|U)). \tag{2}$$

Using the fact that $G$ is $+$, we immediately get

$$\text{Bel}(V \cap (V_1 \cup V_2)|U) = \text{Bel}(V \cap V_1|U) + \text{Bel}(V \cap V_2|U) \tag{3}$$

---

4. I should stress that my counterexample is not a counterexample to Aczél's theorem, since he explicitly assumes that the range of Bel is infinite. However, it does point out potential problems with his proof, and certainly shows that his argument does not apply to finite domains. Aczél is in fact aware of the problems with his proof [private communication, 1996]. He later proved results in a similar spirit with the aid of a requirement of *nonatomicity* (Aczél & Daroczy, 1975, pp. 5–6), which is in fact a stronger requirement than A4, and thus also requires the domain to be infinite.

and

$$F(\text{Bel}(V_1 \cup V_2 | V \cap U), \text{Bel}(V|U))$$
$$= F(\text{Bel}(V_1 | V \cap U) + \text{Bel}(V_2 | V \cap U), \text{Bel}(V|U)) \tag{4}$$

Moreover, by A2, we also have, for $i = 1, 2$,

$$\text{Bel}(V \cap V_i | U) = F(\text{Bel}(V \cap V_i | V \cap U), \text{Bel}(V|U)). \tag{5}$$

Putting together (2), (3), (4), and (5), we get that

$$F(\text{Bel}(V \cap V_1 | V \cap U), \text{Bel}(V|U)) + F(\text{Bel}(V \cap V_2 | V \cap U), \text{Bel}(V|U))$$
$$= F(\text{Bel}(V \cap V_1 | V \cap U) + \text{Bel}(V \cap V_2 | V \cap U), \text{Bel}(V|U)). \tag{6}$$

Taking $x = \text{Bel}(V \cap V_1 | V \cap U)$, $y = \text{Bel}(V \cap V_2 | V \cap U)$, and $z = \text{Bel}(V|U)$ in (6), we get the functional equation

$$F(x, z) + F(y, z) = F(x + y, z). \tag{7}$$

Suppose that we assume (as Reichenbach implicitly does) that this functional equation holds for all $(x, y, z) \in P = \{(x, y, z) \in [0, 1]^3 : x + y \leq 1\}$. The rest of the proof now follows easily. First, taking $x = 0$ in (7), it follows that

$$F(0, z) + F(y, z) = F(y, z),$$

from which we get that

$$F(0, z) = 0.$$

Next, fix $z$ and let $g_z(x) = F(x, z)$. Since $F$ is, by assumption, differentiable, from (7) we have that

$$g_z'(x) = \lim_{y \to 0}(F(x + y, z) - F(x, z)/y) = \lim_{y \to 0} F(y, z)/y.$$

It thus follows that $g_z'(x)$ is a constant, independent of $x$. Since the constant may depend on $z$, there is some function $h$ such that $g_z'(x) = h(z)$. Using the fact that $F(0, z) = 0$, elementary calculus tells us that

$$g_z(x) = F(x, z) = h(z)x.$$

Using the assumption that for all $U, V$, we have $\text{Bel}(V|U) = 1$ if $U \subseteq V$, we get that

$$\text{Bel}(V|U) = \text{Bel}(V \cap V|U) = F(\text{Bel}(V|V \cap U), \text{Bel}(V|U)) = F(1, \text{Bel}(V|U)).$$

Thus, we have that

$$F(1, z) = h(z) = z.$$

We conclude that $F(x, z) = xz$.

Note, however, that this conclusion depends in a crucial way on the assumption that the functional equation (7) holds for all $(x, y, z) \in P$.[5] In fact, all that we can conclude from (6) is that it holds for all $(x, y, z)$ such that there exist $U, V, V_1$, and $V_2$, with $V_1$ and $V_2$ disjoint, such that $x = \text{Bel}(V \cap V_1 | V \cap U)$, $y = \text{Bel}(V \cap V_2 | V \cap U)$, and $z = \text{Bel}(V|U)$.

---

5. Actually, using the continuity of $F$, it suffices that the functional equation holds for a set of triples which is dense in $P$.

Let us say that a triple that satisfies this condition is *R-constrained* (since it must satisfy certain constraints imposed by the $F$ and $G$ functions; the $R$ here is for Reichenbach, to distinguish this notion from a similar one defined in the next section.) As I mentioned earlier, Aczél also assumes that $\mathrm{Bel}(V|U)$ takes on all values in $[e, E]$, where $e = \mathrm{Bel}(\emptyset|U)$ and $E = \mathrm{Bel}(U|U)$. (In Reichenbach's formulation, $e = 0$ and $E = 1$.) There are two ways to interpret this assumption. The weak interpretation is that for each $x \in [0, 1]$, there exist $U, V$ such that $\mathrm{Bel}(V|U) = x$. The strong interpretation is that for each $U$ and $x$, there exists $V$ such that $\mathrm{Bel}(V|U) = x$. It is not clear which interpretation is intended by Aczél. Neither one obviously suffices to prove that every triple in $P$ is R-constrained, although it does seem plausible that it might follow from the second assumption.

In any case, neither Aczél nor Reichenbach see a need to check that Equation (7) holds throughout $P$. (Nor does Cox for his analogous functional equation, nor do the authors of more recent and polished presentations of Cox's result, such as Jaynes (1996) and Tribus (1969).) However, it turns out to be quite necessary to do this. Moreover, it is clear that if $W$ is finite, there are only finitely tuples in $P$ that are R-constrained, and it is not the case that all of $P$ is. As we shall see in the next section, this observation has serious consequences as far as all these proofs are concerned.

## 3. The Counterexample to Cox's Theorem

The goal of this section is to prove

**Theorem 3.1:** *There is a function $\mathrm{Bel}_0$, a finite domain $W$, and functions $S$, $F$, and $G$ satisfying A1, A2, and A3 respectively such that*

- *$\mathrm{Bel}_0(V|U) \in [0, 1]$ for $U \neq \emptyset$,*

- *$S(x) = 1 - x$ (so that $S$ is strictly decreasing and infinitely differentiable),*

- *$G(x, y) = x + y$ (so that $G$ is strictly increasing in each argument and is infinitely differentiable),*

- *$F$ is infinitely differentiable, nondecreasing in each argument in $[0, 1]^2$, and strictly increasing in each argument in $(0, 1]^2$. Moreover, $F$ is commutative, $F(x, 0) = F(0, x) = 0$, and $F(x, 1) = F(1, x) = x$.*

*However, there is no one-to-one onto function $g : [0, 1] \to [0, 1]$ satisfying (1).*

Note that the hypotheses on $\mathrm{Bel}_0$, $S$, $G$, and $F$ are at least as strong as those made in all the other variants of Cox's result, while the assumptions on $g$ are weaker than those made in the variants. For example, there is no requirement that $g$ be continuous or increasing nor that $g \circ \mathrm{Bel}_0$ is a probability distribution (although Paris and Aczél both prove that, under their assumptions, $g$ can be taken to satisfy all these requirements). This serves to make the counterexample quite strong.

The proof of Theorem 3.1 is constructive. Consider a domain $W$ with 12 points: $w_1, ..., w_{12}$. We associate with each point $w \in W$ a weight $f(w)$, as follows.

$$
\begin{array}{ll}
f(w_1) = 3 & f(w_4) = 5 \times 10^4 \\
f(w_2) = 2 & f(w_5) = 6 \times 10^4 \\
f(w_3) = 6 & f(w_6) = 8 \times 10^4
\end{array}
$$

$$
\begin{array}{ll}
f(w_7) = 3 \times 10^8 & f(w_{10}) = 3 \times 10^{18} \\
f(w_8) = 8 \times 10^8 & f(w_{11}) = 2 \times 10^{18} \\
f(w_9) = 8 \times 10^8 & f(w_{12}) = 14 \times 10^{18}
\end{array}
$$

For a subset $U$ of $W$, we define $f(U) = \sum_{w \in U} f(w)$. Thus, we can define a probability distribution $\Pr$ on $W$ by taking $\Pr(U) = f(U)/f(W)$.

Let $f'$ be identical to $f$, except that $f'(w_{10}) = (3 - \delta) \times 10^{18}$ and $f'(w_{11}) = (2 + \delta) \times 10^{18}$, where $\delta$ is defined below. Again, we extend $f'$ to subsets of $W$ by defining $f'(U) = \sum_{w \in U} f'(w)$. Let $W' = \{w_{10}, w_{11}, w_{12}\}$. If $U \neq \emptyset$, define

$$
\mathrm{Bel}_0(V|U) = \left\{ \begin{array}{ll} f'(V \cap U)/f'(U) & \text{if } W' \subseteq U \\ f(V \cap U)/f(U) & \text{otherwise.} \end{array} \right.
$$

$\mathrm{Bel}_0$ is clearly very close to $\Pr$. If $U \neq \emptyset$, then it is easy to see that $|\mathrm{Bel}_0(V|U) - \Pr(V|U)| = |f'(V \cap U) - f(V \cap U)|/f(U) \leq \delta$. We choose $\delta > 0$ so that

$$\text{if } \Pr(V|U) > \Pr(V'|U'), \text{ then } \mathrm{Bel}_0(V|U) > \mathrm{Bel}_0(V'|U'). \tag{8}$$

Since the range of $\Pr$ is finite, all sufficiently small $\delta$ satisfy (8).

The exact choice of weights above is not particularly important. One thing that is important though is the following collection of equalities:

$$
\begin{array}{l}
\Pr(w_1|\{w_1, w_2\}) = \Pr(w_{10}|\{w_{10}, w_{11}\}) = 3/5 \\
\Pr(\{w_1, w_2\}|\{w_1, w_2, w_3\}) = \Pr(w_4|\{w_4, w_5\}) = 5/11 \\
\Pr(\{w_4, w_5\}|\{w_4, w_5, w_6\}) = \Pr(\{w_7, w_8\}|\{w_7, w_8, w_9\}) = 11/19 \\
\Pr(w_4|\{w_4, w_5, w_6\}) = \Pr(\{w_{10}, w_{11}\}|\{w_{10}, w_{11}, w_{12}\}) = 5/19 \\
\Pr(w_1|\{w_1, w_2, w_3\}) = \Pr(w_7|\{w_7, w_8\}) = 3/11.
\end{array} \tag{9}
$$

It is easy to check that exactly the same equalities hold if we replace $\Pr$ by $\mathrm{Bel}_0$.

We show that $\mathrm{Bel}_0$ satisfies the requirements of Theorem 3.1 by a sequence of lemmas. The first lemma is the key to showing that $\mathrm{Bel}_0$ cannot be isomorphic to a probability function. It uses the fact (proved in Lemma 3.3) that if $\mathrm{Bel}_0$ were isomorphic to a probability function, then there would have to be a function $F$ satisfying A2 that is associative. Although, as is shown in Lemma 3.7, the function $F$ satisfying A2 can be taken to be infinitely differentiable and increasing in each argument, the equalities in (9) suffice to guarantee that it cannot be taken to be associative, that is, we do not in general have

$$F(x, F(y, z)) = F(F(x, y), z).$$

Indeed, there is no associative function $F$ satisfying A2, even if we drop the requirements that $F$ be differentiable or increasing.

**Lemma 3.2:** *For $Bel_0$ as defined above, there is no associative function $F$ satisfying A2.*

**Proof:** Suppose there were such a function $F$. From (9), we must have that

$$
\begin{aligned}
&F(5/11, 11/19)\\
=\ &F(\mathrm{Bel}_0(w_4|\{w_4, w_5\}), \mathrm{Bel}_0(\{w_4, w_5\}|\{w_4, w_5, w_6\}))\\
=\ &\mathrm{Bel}_0(w_4|\{w_4, w_5, w_6\}) = 5/19
\end{aligned}
$$

and that

$$
\begin{aligned}
&F(3/5, 5/11)\\
=\ &F(\mathrm{Bel}_0(w_1|\{w_1, w_2\}), \mathrm{Bel}_0(\{w_1, w_2\}|\{w_1, w_2, w_3\}))\\
=\ &\mathrm{Bel}_0(w_1|\{w_1, w_2, w_3\}) = 3/11.
\end{aligned}
$$

It follows that

$$
F(3/5, F(5/11, 11/19)) = F(3/5, 5/19)
$$

and that

$$
F(F(3/5, 5/11), 11/19) = F(3/11, 11/19).
$$

Thus, if $F$ were associative, we would have

$$
F(3/5, 5/19) = F(3/11, 11/19).
$$

On the other hand, from (9) again, we see that

$$
\begin{aligned}
&F(3/5, 5/19)\\
=\ &F(\mathrm{Bel}_0(w_{10}|\{w_{10}, w_{11}\}), \mathrm{Bel}_0(\{w_{10}, w_{11}\}|\{w_{10}, w_{11}, w_{12}\}))\\
=\ &\mathrm{Bel}_0(w_{10}|\{w_{10}, w_{11}, w_{12}\}) = (3-\delta)/19,
\end{aligned}
$$

while

$$
\begin{aligned}
&F(3/11, 11/19)\\
=\ &F(\mathrm{Bel}_0(w_7|\{w_7, w_8\}), \mathrm{Bel}_0(\{w_7, w_8\}|\{w_7, w_8, w_9\}))\\
=\ &\mathrm{Bel}_0(w_7|\{w_7, w_8, w_9\}) = 3/19.
\end{aligned}
$$

It follows that $F$ cannot be associative. $\square$

To understand how Lemma 3.2 relates to our discussion in Section 2 of the problems with Reichenbach's proof, we say $(x, y, z)$ is a *constrained triple* if there exist sets $U_1 \supseteq U_2 \supseteq U_3 \supseteq U_4$ with $U_3 \neq \emptyset$ such that $x = \mathrm{Bel}_0(U_4|U_3)$, $y = \mathrm{Bel}_0(U_3|U_2)$, and $z = \mathrm{Bel}_0(U_2|U_1)$. It is easy to see that A2 forces $F$ to be associative on constrained triples, since if $w = \mathrm{Bel}_0(U_3|U_1)$ and $w' = \mathrm{Bel}_0(U_4|U_2)$, by A2, we have $F(x, F(y, z)) = F(x, w) = \mathrm{Bel}_0(U_4|U_1)$ and $F(F(x, y), z) = F(w', z) = \mathrm{Bel}_0(U_4, U_1)$. A4 says that the set of constrained triples is dense in $[0, 1]^3$.

We similarly define $(x, y)$ to be a *constrained pair* if there exist sets $U_1 \supseteq U_2 \supseteq U_3$ with $U_2 \neq \emptyset$ such that $x = \mathrm{Bel}_0(U_3|U_2)$ and $y = \mathrm{Bel}_0(U_2|U_1)$. We say that $(U_1, U_2, U_3)$ *corresponds* to the constrained pair $(x, y)$. (Note that there may be more than one triple of sets corresponding to a constrained pair.) If $(U_1, U_2, U_3)$ corresponds to the constrained pair $(x, y)$ and $F$ satisfies A2, then we must have $F(x, y) = \mathrm{Bel}_0(U_3|U_1)$. Note that both $(3/5, 5/11)$ and $(5/11, 11/19)$ are constrained pairs, although the triple $(3/5, 5/11, 11/19)$ is not constrained. It is this fact that we use in Lemma 3.2.

The next lemma shows that $\mathrm{Bel}_0$ cannot be isomorphic to a probability function.

**Lemma 3.3:** *For $Bel_0$ as defined above, there is no one-to-one onto function $g : [0, 1] \rightarrow [0, 1]$ satisfying (1).*

**Proof:** Suppose there were such a function $g$. First note that $g(\text{Bel}_0(U)) \neq 0$ if $U \neq \emptyset$. For if $g(\text{Bel}_0(U)) = 0$, then it follows from (1) that for all $V \subseteq U$, we have

$$g(\text{Bel}_0(V)) = g(\text{Bel}_0(V|U)) \times g(\text{Bel}_0(U)) = g(\text{Bel}_0(V|U)) \times 0 = 0.$$

Thus, $g(\text{Bel}_0(V)) = g(\text{Bel}_0(U))$ for all subsets $V$ of $U$. Since the definition of $\text{Bel}_0$ guarantees that $\text{Bel}_0(V) \neq \text{Bel}_0(U)$ if $V$ is a strict subset of $U$, this contradicts the assumption that $g$ is one-to-one. Thus, $g(\text{Bel}_0(U)) \neq 0$ if $U \neq \emptyset$. It now follows from (1) that if $U \neq \emptyset$, then

$$g(\text{Bel}_0(V|U)) = g(\text{Bel}_0(V \cap U))/g(\text{Bel}_0(U)). \tag{10}$$

Now define $F(x, y) = g^{-1}(g(x) \times g(y))$. We show that $F$ defined in this way satisfies A2 and is associative. This will give us a contradiction to Lemma 3.2.

To see that $F$ satisfies A2, notice that, by applying the observation above repeatedly, if $V \cap U \neq \emptyset$, we get

$$
\begin{aligned}
&F(\text{Bel}_0(V'|V \cap U), \text{Bel}_0(V|U))\\
=\ & g^{-1}((g(\text{Bel}_0(V'|V \cap U)) \times g(\text{Bel}_0(V|U)))\\
=\ & g^{-1}((g(\text{Bel}_0(V' \cap V \cap U))/g(\text{Bel}_0(V \cap U))) \times (g(\text{Bel}_0(V \cap U))/g(\text{Bel}_0(U))))\\
=\ & g^{-1}(g(\text{Bel}_0(V' \cap V \cap U))/g(\text{Bel}_0(U)))\\
=\ & g^{-1}(g(\text{Bel}_0(V' \cap V|U)))\\
=\ & \text{Bel}_0(V' \cap V|U).
\end{aligned}
$$

Thus, $F$ satisfies A2.

To see that $F$ is associative, note that

$$
\begin{aligned}
F(F(x, y), z) &= g^{-1}(g(g^{-1}(g(x) \times g(y))) \times g(z))\\
&= g^{-1}(g(x) \times g(y) \times g(z))\\
&= g^{-1}(g(x) \times g(g^{-1}(g(y) \times g(z))))\\
&= F(x, F(y, z)).
\end{aligned}
$$

This gives us the desired contradiction to Lemma 3.2. It follows that $\text{Bel}_0$ cannot be isomorphic to a probability function. $\square$

Despite the fact that $\text{Bel}_0$ is not isomorphic to a probability function, functions $S$, $F$, and $G$ can be defined that satisfy A1, A2, and A3, respectively, and all the other requirements stated in Theorem 3.1. The argument for $S$ and $G$ is easy; all the work goes into proving that an appropriate $F$ exists.

**Lemma 3.4:** *There exists an infinitely differentiable, strictly decreasing function $S : [0, 1] \rightarrow [0, 1]$ such that $Bel_0(\overline{V}|U) = S(Bel_0(V|U))$ for all sets $U, V \subseteq W$ with $U \neq \emptyset$. In fact, we can take $S(x) = 1 - x$.*

**Proof:** This is immediate from the observation that $\text{Bel}_0(\overline{V}|U) = 1 - \text{Bel}_0(V|U)$ for $U, V \subseteq W$. $\square$

**Lemma 3.5:** *There exists an infinitely differentiable function* $G : [0,1]^2 \to [0,1]$, *increasing in each argument, such that if* $U, V, V' \subseteq W$, $V \cap V' = \emptyset$, *and* $U \neq \emptyset$, *then* $\mathrm{Bel}_0(V \cup V'|U) = G(\mathrm{Bel}_0(V|U), \mathrm{Bel}_0(V',U))$. *In fact, we can take* $G(x,y) = x + y$.

**Proof:** This is immediate from the definition of $\mathrm{Bel}_0$. □

Thus, all that remains is to show that an appropriate $F$ exists. The key step is provided by the following lemma, which essentially shows that there is a well defined $F$ that is increasing.

**Lemma 3.6:** *If* $U_2 \cap U_1 \neq \emptyset$ *and* $V_2 \cap V_1 \neq \emptyset$, *then*

(a) *if* $\mathrm{Bel}_0(V_3|V_2 \cap V_1) \leq \mathrm{Bel}_0(U_3|U_2 \cap U_1)$ *and* $\mathrm{Bel}_0(V_2|V_1) \leq \mathrm{Bel}_0(U_2|U_1)$, *then* $\mathrm{Bel}_0(V_3 \cap V_2|V_1) \leq \mathrm{Bel}_0(U_3 \cap U_2|U_1)$,

(b) *if* $\mathrm{Bel}_0(V_3|V_2 \cap V_1) < \mathrm{Bel}_0(U_3|U_2 \cap U_1)$, $\mathrm{Bel}_0(V_2|V_1) \leq \mathrm{Bel}_0(U_2|U_1)$, $\mathrm{Bel}_0(U_3|U_2 \cap U_1) > 0$, *and* $\mathrm{Bel}_0(U_2|U_1) > 0$, *then* $\mathrm{Bel}_0(V_3 \cap V_2|V_1) < \mathrm{Bel}_0(U_3 \cap U_2|U_1)$,

(c) *if* $\mathrm{Bel}_0(V_3|V_2 \cap V_1) \leq \mathrm{Bel}_0(U_3|U_2 \cap U_1)$, $\mathrm{Bel}_0(V_2|V_1) < \mathrm{Bel}_0(U_2|U_1)$, $\mathrm{Bel}_0(U_3|U_2 \cap U_1) > 0$, *and* $\mathrm{Bel}_0(U_2|U_1) > 0$, *then* $\mathrm{Bel}_0(V_3 \cap V_2|V_1) < \mathrm{Bel}_0(U_3 \cap U_2|U_1)$,

**Proof:** First observe that if $\mathrm{Bel}_0(V_3|V_2 \cap V_1) \leq \mathrm{Bel}_0(U_3|U_2 \cap U_1)$ and $\mathrm{Bel}_0(V_2|V_1) \leq \mathrm{Bel}_0(U_2|U_1)$, then from (8), it follows that $\mathrm{Pr}(V_3|V_2 \cap V_1) \leq \mathrm{Pr}(U_3|U_2 \cap U_1)$ and $\mathrm{Pr}(V_2|V_1) \leq \mathrm{Pr}(U_2|U_1)$. If we have either $\mathrm{Pr}(V_3|V_2 \cap V_1) < \mathrm{Pr}(U_3|U_2 \cap U_1)$ or $\mathrm{Pr}(V_2|V_1) < \mathrm{Pr}(U_2|U_1)$, then we have either $\mathrm{Pr}(V_3 \cap V_2|V_1) < \mathrm{Pr}(U_3 \cap U_2|U_1)$ or $\mathrm{Pr}(U_3|U_2 \cap U_1) = 0$ or $\mathrm{Pr}(U_2|U_1) = 0$. It follows that either $\mathrm{Bel}_0(V_3 \cap V_2|V_1) < \mathrm{Bel}_0(U_3 \cap U_2|U_1)$ (this uses (8) again) or that $\mathrm{Bel}_0(V_3 \cap V_2|V_1) = \mathrm{Bel}_0(U_3 \cap U_2|U_1) = 0$. In either case, the lemma holds.

Thus, it remains to deal with the case that $\mathrm{Pr}(V_3|V_2 \cap V_1) = \mathrm{Pr}(U_3|U_2 \cap U_1)$ and $\mathrm{Pr}(V_2|V_1) = \mathrm{Pr}(U_2|U_1)$, and hence $\mathrm{Pr}(V_3 \cap V_2|V_1) = \mathrm{Pr}(U_3 \cap U_2|U_1)$. The details of this analysis are left to the appendix. □

**Lemma 3.7:** *There exists a function* $F : [0,1]^2 \to [0,1]$ *satisfying all the assumptions of Theorem 3.1 (with respect to* $\mathrm{Bel}_0$).

**Proof:** Define a partial function $F'$ on $[0,1]^2$ whose domain $D$ consists of all constrained pairs. For a constrained pair, we define $F'$ in the unique way required to satisfy A2. *A priori*, $F'$ may not be well defined; it is possible that there exist triples $(U_1, U_2, U_3)$ and $(V_1, V_2, V_3)$ that both correspond to $(x, y)$ (i.e., $x = \mathrm{Bel}_0(U_3|U_2) = \mathrm{Bel}_0(V_3|V_2)$ and $y = \mathrm{Bel}_0(U_2|U_1) = \mathrm{Bel}_0(V_2|V_1)$) such that $\mathrm{Bel}_0(U_3|U_1) \neq \mathrm{Bel}_0(V_3|V_1)$. If this were the case, then $F'(x, y)$ would not be well defined. However, Lemma 3.6 says that this cannot happen. Moreover, Lemma 3.6 assures us that $F'$ is increasing on $D$, and strictly increasing as long as one of its arguments is not 0. Indeed, if there is a triple $(U_1, U_2, U_3)$ corresponding to $(x, y)$ such that $\{w_{10}, w_{11}, w_{12}\} \not\subseteq U_1$, then we must have $F'(x, y) = xy$.

The domain $D$ of $F'$ is finite. Let $D'$ be the commutative closure of $D$, so that $D'$ consists of $D$ and all pairs $(y, x)$ such that $(x, y)$ is in $D$. Extend $F'$ to a commutative function $F''$ on $D'$ by defining $F''(y, x) = F'(x, y)$ if $(x, y) \in D$. $F''$ is well defined because, as can easily be verified, if $(x, y)$ and $(y, x)$ are both in $D$, one of $x$ or $y$ must be 1, and $F'(x, 1) = F'(1, x) = x$. Clearly $F''$ is commutative. It is also increasing. For suppose $(x, y), (x', y') \in D'$, $x \leq x'$, and $y \leq y'$. If both $(x, y)$ and $(x', y')$ are in $D$, we must have $F''(x, y) \leq F''(x', y')$, since $F'$ is increasing. Similarly, if both $(y, x)$ and $(y', x')$ are in $D$, we must have $F''(x, y) = F'(y, x) \leq F'(y', x') = F''(x', y')$. Finally, if $(x, y)$ and $(y', x')$ are in $D$, a straightforward check over all possible elements in $D$ shows that this can happen only if the triples $(U_1, U_2, U_3)$ and $(V_1, V_2, V_3)$ corresponding to $(x, y)$ and $(y', x')$ are such that $\{w_{10}, w_{11}, w_{12}\}$ is not a subset of either $U_1$ or $V_1$. It follows that $F'(x, y) = xy$ and $F'(y', x') = x'y'$, so again we get that $F''$ is increasing. A similar argument shows that $F''$ is strictly increasing as long as one of its arguments is not 0.

It is straightforward to extend $F''$ to a commutative, infinitely differentiable, and increasing function $F$ defined on all of $[0, 1]^2$, which is strictly increasing on $(0, 1)^2$, and satisfies $F(x, 1) = F(1, x) = x$ and $F(x, 0) = F(0, x) = 0$. We proceed as follows. We first extend $F''$ so that it is defined for all pairs $(x, y) \in [0, 1]^2$ such that $x \geq y$ so that it has the required properties. If $x < y$, we then define so that $F(x, y) = F(y, x)$. Since $F''$ is commutative, this definition agrees with $F''(x, y)$ for $x < y$. Clearly $F$ is commutative and infinitely differentiable. To see that $F$ is increasing, suppose that $x \leq x'$ and $y \leq y'$. Just as in the case of $F''$, it is immediate that $F$ is increasing if both $x \geq y$ and $x' \geq y'$ or both $x < y$ and $x' < y'$. Otherwise, suppose $x \geq y$ and $y' \geq x'$. Then we have $y \leq x \leq x' \leq y'$. Since $F$ is increasing on $\{(x, y) : x \geq y\}$, we have $F(x, y) \leq F(x', y) \leq F(x', x') \leq F(y', x') = F(x', y')$. A similar argument shows that $F$ is strictly increasing unless one its arguments is 0. Finally, $F$ clearly satisfies A2, since (by construction) $F'$ does, and A2 puts constraints only on the domain of $F'$. $\square$

Theorem 3.1 now follows from Lemmas 3.3, 3.4, 3.5, and 3.7.

## 4. The Counterexample to Fine's Theorem

Fine is interested in what he calls *comparative conditional probability*. Thus, rather than associating a real number with each "conditional object" $V|U$, he puts an ordering $\succeq$ on such objects. As usual, $V|U \succ V'|U'$ is taken to be an abbreviation for $V|U \succeq V'|U'$ and $\text{not}(V'|U' \succeq V|U)$.

Fine is interested in when such an ordering is induced by a real-valued belief function with reasonable properties. He says that a real-valued function $P$ on such objects *agrees with* $\succeq$ if $P(V|U) \geq P(V'|U')$ iff $V|U \succeq V'|U'$. Fine then considers a number of axioms that $\succeq$ might satisfy. For our purposes, the most relevant are the ones Fine denotes QCC1, QCC2, QCC5, and QCC7.

QCC1 just says that $\succeq$ is a linear order:

**QCC1.** $V|U \succeq V'|U'$ or $V'|U' \succeq V|U$.

QCC2 says that $\succeq$ is transitive:

**QCC2.** If $V_1|U_1 \succeq V_2|U_2$ and $V_2|U_2 \succeq V_3|U_3$, then $V_1|U_1 \succeq V_3|U_3$.

QCC5 is a technical condition involving notions of order topology. The relevant definitions are omitted here (see (Fine, 1973) for details), since QCC5, as Fine observes, holds vacuously in finite domains (the only ones of interest here).

**QCC5.** The set $\{V|U\}$ has a countable basis in the order topology induced by $\succ$.

Finally, QCC7 essentially says that $\succeq$ is increasing, in the sense of Lemma 3.6.

**QCC7.**

    (a) If $V_3|V_2 \cap V_1 \succeq U_3|U_2 \cap U_1$ and $V_2|V_1 \succeq U_2|U_1$ then $V_3 \cap V_2|V_1 \succeq U_3 \cap U_2|U_1$.

    (b) If $V_3|V_2 \cap V_1 \succeq U_2|U_1$ and $V_2|V_1 \succeq U_3|U_2 \cap U_1$ then $V_3 \cap V_2|V_1 \succeq U_3 \cap U_2|U_1$.

    (c) If $V_3|V_2 \cap V_1 \succ U_3|U_2 \cap U_1$, $V_2|V_1 \succeq U_2|U_1$, and $V_2|V_1 \succ \emptyset|W$, then $V_3 \cap V_2|V_1 \succ U_3 \cap U_2|U_1$.

Fine then claims the following theorem:

**Fine's Theorem:** (Fine, 1973, Chapter II, Theorem 8) *If $\succeq$ satisfies QCC1, QCC2, QCC5, then there exists some agreeing function $P$. There exists a function $F$ of two variables such that*

    *1. $P(V \cap V'|U) = F(P(V'|V \cap U), P(V|U))$,[6]*

    *2. $F(x, y) = F(y, x)$,*

    *3. $F(x, y)$ is increasing in $x$ for $y > P(\emptyset|W)$,*

    *4. $F(x, F(y, z)) = F(F(x, y), z)$,*

    *5. $F(P(W|U), y) = y$,*

    *6. $F(P(\emptyset|U), y) = P(\emptyset|U)$.*

*iff $\succeq$ also satisfies QCC7.*

The only relevant clauses for our purposes are Clause (1), which is just A2, and Clause (4), which says that $F$ is associative. As Lemma 3.2 shows, there is no associative function satisfying A2 for $\text{Bel}_0$. As I now show, this means that Fine's theorem does not quite hold either.

Before doing so, let me briefly touch on a subtle issue regarding the domain of $\succeq$. In the counterexample of the previous section, $\text{Bel}_0(V|U)$ is defined as long as $U \neq \emptyset$. Fine does not assume that the $\succeq$ relation is necessarily defined on all objects $V|U$ such that $U, V \subseteq W$ and $U \neq \emptyset$. He assumes that there is an algebra $\mathcal{F}$ of subsets of $W$ (that is, a set of subsets closed under finite intersections and complementation) and a subset $\mathcal{F}'$ of $\mathcal{F}$ closed under finite intersections and not containing the empty set such that $\succeq$ is defined on conditional objects $V|U$ such that $V \in \mathcal{F}$ and $U \in \mathcal{F}'$. Since $\mathcal{F}'$ is closed under intersection and does not contain the empty set, $\mathcal{F}'$ cannot contain disjoint sets. If $W$ is finite, then the only way a collection $\mathcal{F}'$ can meet Fine's restriction is if there is some nonempty set $U_0$ such that all elements in $\mathcal{F}'$ contain $U_0$. This restriction is clearly too strong to the extent that comparative conditional probability is intended to generalize probability. If $\text{Pr}$ is a probability function, then it certainly makes sense to compare $\text{Pr}(V|U)$ and $\text{Pr}(V'|U')$ even if $U$ and $U'$ are disjoint sets. Fine [private communication, 1995] suggested that it might be better to constrain QCC7 so that we do not condition on events $U$ that are equivalent to $\emptyset$ (where $U$ is equivalent to $\emptyset$ if $\emptyset \succeq U$ and $U \succeq \emptyset$). Since the only event equivalent to $\emptyset$ in the counterexample of the previous section is $\emptyset$ itself, this means that the counterexample can be used without change. This is what is done in the proof below. I show below how to modify the counterexample so that it satisfies Fine's original restrictions.

---

6. Fine assumes that $P(V \cap V'|U) = F(P(V|U), P(V'|V \cap U))$. I have reordered the arguments here for consistency with Cox's theorem.

**Theorem 4.1:** *There exists an ordering $\succeq$ satisfying QCC1, QCC2, QCC5, and QCC7, such that for every function $P$ agreeing with $\succeq$, there is no associative function $F$ of two variables such that $P(V \cap V')|U) = F(P(V'|V \cap U), P(V|U))$.*

**Proof:** Let $W$ and $\mathrm{Bel}_0$ be as in the counterexample in the previous section. Define $\succeq$ so that $\mathrm{Bel}_0$ agrees with $\succeq$. Thus, $V|U \succeq V'|U'$ iff $\mathrm{Bel}_0(V|U) \geq \mathrm{Bel}_0(V'|U')$. Clearly $\succeq$ satisfies QCC1 and QCC2. As was mentioned earlier, since $W$ is finite, $\succeq$ vacuously satisfies QCC5. Lemma 3.6 shows that $\succeq$ satisfies parts (a) and (c) of QCC7. To show that $\succeq$ also satisfies part (b) of QCC7, we must prove that if $\mathrm{Bel}_0(V_3|V_2 \cap V_1) \geq \mathrm{Bel}_0(U_2|U_1)$ and $\mathrm{Bel}_0(V_2|V_1) \geq \mathrm{Bel}_0(U_3|U_2 \cap U_1)$, then $\mathrm{Bel}_0(V_3 \cap V_2|V_1) \geq \mathrm{Bel}_0(U_3 \cap U_2|U_1)$. The proof of this is almost identical to that of Lemma 3.6; we simply exchange the roles of $\mathrm{Pr}(V_2|V_1)$ and $\mathrm{Pr}(V_3|V_2 \cap V_1)$ in that proof. I leave the details to the reader. Lemma 3.2 shows that there is no associative function $F$ satisfying A2 for $\mathrm{Bel}_0$. All that was used in the proof was the fact that $\mathrm{Bel}_0$ satisfied the inequalities of (9). But these equalities must hold for any function agreeing with $\succeq$. Thus, exactly the same proof shows that if $P$ is any function agreeing with $\succeq$, then there is no associative function $F$ satisfying $P(V \cap V'|U) = F(P(V'|V \cap U), P(V|U))$. □

I conclude this section by briefly sketching how the counterexample can be modified so that it satisfies Fine's original restriction. Redefine $W$ by adding one more element $w_0$. Redefine $f$ and $f'$ so that $f(w_0) = f'(w_0) = 10^{-5}$; in addition, redefine $f$ and $f'$ on $w_3$, $w_6$, $w_9$, and $w_{12}$, so as to decrease their weight by $10^{-5}$, the weight of $w_0$. Thus,

- $f(w_3) = f'(w_3) = 6 - 10^{-5}$,

- $f(w_6) = f'(w_6) = 8 \times 10^4 - 10^{-5}$,

- $f(w_9) = f'(w_9) = 8 \times 10^8 - 10^{-5}$, and

- $f(w_{12}) = f'(w_{12}) = 14 \times 10^{18} - 10^{-5}$.

Finally, redefine $W'$ to be $\{w_0, w_{10}, w_{11}, w_{12}\}$. The definition of $\mathrm{Bel}_0$ in terms of $f$, $f'$, and $W'$ remains the same. With these redefinitions, the proofs of the previous section go through essentially unchanged. In particular, the equalities in (9) now hold if we add $w_0$ to every set. Let $\mathcal{F}'$ consist of all subsets of $W$ containing $w_0$. Notice that $\mathcal{F}'$ is closed under intersection and does not contain the empty set. The lack of associativity in Lemma 3.2 can now be demonstrated by conditioning on sets in $\mathcal{F}'$. As a consequence, we get a counterexample to Fine's theorem even when restricting to conditional objects that satisfy his restriction.

## 5. Discussion

Let me summarize the status of various results in the light of the counterexample of this paper:

- Cox's theorem as originally stated does not hold in finite domains. Moreover, even in infinite domains, the counterexample and the discussion in Section 2 suggest that more assumptions are required for its correctness. In particular, the claim in his proof that $F$ is associative does not follow.

- Although the counterexample given here is not a counterexample to Aczél's theorem, his assumptions do not seem strong enough to guarantee that the function $G$ is associative, as he claims it is.

- The variants of Cox's theorem stated by Heckerman (1988), Horvitz, Heckerman, and Langlotz (1986), and Aleliunas (1988) all succumb to the counterexample.

- The claim that the function $F$ must be associative in Fine's theorem is incorrect. Fine has an analogous result (Fine, 1973, Chapter II, Theorem 4) for unconditional comparative probability involving a function $G$ as in Aczél's theorem. This function too is claimed to be associative, and again, this does not seem to follow (although my counterexample does not apply to that theorem).

Of course, the interesting question now is what it would take to recover Cox's theorem. Paris's assumption A4 suffices, as does the stronger assumption of nonatomicity (see Footnote 4). As we have observed, A4 forces the domain of Bel to be infinite, as does the assumption that the range of Bel is all of $[0, 1]$. We can always extend a domain to an infinite—indeed, uncountable—domain by assuming that we have an infinite collection of independent fair coins, and that we can talk about outcomes of coin tosses as well as the original events in the domain. (This type of "extendibility" assumption is fairly standard; for example, it is made by Savage (1954) in quite a different context.) In such an extended domain, it seems reasonable to also assume that Bel varies uniformly between 0 (certain falsehood) and 1 (certain truth). If we also assume A4 (or something like it), we can then recover Cox's theorem. Notice, however, that this viewpoint disallows a notion of belief that takes on only finitely many gradations.

Another possibility is to observe that we are not interested in just one domain in isolation. Rather, what we are interested in is a notion of belief Bel that applies uniformly to all domains. Thus, even if $(U, V)$ and $(U', V')$ are pairs of subsets of different (perhaps even disjoint) domains, if $\mathrm{Bel}(V|U)$ and $\mathrm{Bel}(V'|U')$ are both $1/2$, then we would expect this to denote the same relative strength of belief. In this setting, an analogue of A4 seems more reasonable. That is, we can assume that for all $0 \leq \alpha, \beta, \gamma \leq 1$ and $\epsilon > 0$, there is some domain $W$ and subsets $U_1$, $U_2$, $U_3$, and $U_4$ of $W$ such that the conclusion of A4 holds. If we further assume that the functions $F$, $G$, and $S$ are also uniform across domains (that is, that A1, A2, and A3 hold for the same choice of $F$, $G$, and $S$ in every domain), then we can again recover Cox's theorem.[7]

---

7. This point was independently observed by Jeff Paris [private communication, 1996].

The idea of having a notion of uncertainty that applies uniformly in all domains seems implicit in some discussion in that Jaynes' recent book on probability theory (1996). Jaynes focuses almost exclusively on finite domains.[8] As he says "In principle, every problem must start with such finite set probabilities; extensions to infinite sets is permitted only when this is the result of a well-defined and well-behaved limiting process from a finite set." To make sense of this limiting process, it seems that Jaynes must be assuming that the same notion of uncertainty applies in all domains. Moreover, one can make arguments appealing to continuity that when we consider such limiting processes, we can always find subsets $U_1$, $U_2$, $U_3$, and $U_4$ in some sufficiently rich (but finite) extension of the original domain such that A4 holds.

While this seems like perhaps the most reasonable additional assumptions required to get Cox's result, it does require us to consider many domains at once. Moreover, it does not allow a notion of belief that has only finitely many gradations, let alone a notion of belief that allows some events to be considered incomparable in likelihood.[9]

Suppose we really are interested in one particular finite domain, and we do not want to extend it or consider all other possible domains. What assumptions do we then need to get Cox's theorem? The counterexample given here could be circumvented by requiring that $F$ be associative on all tuples (rather than just on the constrained triples). However, if we really are interested in a single domain, the motivation for making requirements on the behavior of $F$ on belief values that do not arise is not so clear. Moreover, it is far from clear that assuming that $F$ is associative suffices to prove the theorem. For example, Cox's proof makes use of various functional equations involving $F$ and $S$, analogous to the equation (7) that appears in Section 2. These functional equations are easily seen to hold for certain tuples. However, as we saw in Section 2, the proof really requires that they hold for *all* tuples. Just assuming that $F$ is associative does not appear to suffice to guarantee that the functional equations involving $S$ hold for all tuples. Further assumptions appear necessary.

Nir Friedman [private communication] has conjectured that the following condition, which says that essentially all beliefs are distinct, suffices:

- if $\emptyset \subset U \subset V$, $\emptyset \subset U' \subset V'$, and $(U, V) \neq (U', V')$, then $\text{Bel}(U|V) \neq \text{Bel}(U'|V')$.

Even if this condition suffices, note that it precludes, for example, a uniform probability distribution, and thus again seems unduly restrictive.

Another possibly interesting line of research is that of characterizing the functions that satisfy Cox's assumptions. As the example given here shows, the class of such functions includes functions that are not isomorphic to any probability function. I conjecture that in fact it includes only functions that are in some sense "close" to a function isomorphic to a probability distribution, although it is not clear exactly how "close" should be defined (nor how interesting this class really is in practice).

So what does all this say regarding the use of probability? Not much. Although I have tried to argue here that Cox's justification of probability is not quite as strong as previously believed, and the assumptions underlying the variants of it need clarification, I am not trying to suggest that probability should be abandoned. There are many other justifications for its use.

---

8. Actually, Jaynes assigns probability to propositions, not sets, but, as noted earlier, there is essentially no difference between the two.

9. Interestingly, Jaynes (1996, Appendix A) admits that having plausibility values be elements of a partially-ordered lattice may be a reasonable alternative to traditional probability theory. Nir Friedman and I (1995, 1996, 1997) have recently developed such a theory and shown that it provides a useful basis for thinking about default reasoning and belief revision.

## Acknowledgments

I'd like to thank Janos Aczél, Peter Cheeseman, Terry Fine, Ron Fagin, Nir Friedman, David Heckerman, Eric Horvitz, Christopher Meek, Jeff Paris, and the anonymous referees for useful comments on the paper. I'd also like to thank Judea Pearl for pointing out Reichenbach's work to me and Janos Aczél for pointing out Falmagne's paper. This work was largely carried out while I was at the IBM Almaden Research Center. IBM's support is gratefully acknowledged. The work was also supported in part by the NSF, under grants IRI-95-03109 and IRI-96-25901, and the Air Force Office of Scientific Research (AFSC), under grant F94620-96-1-0323. A preliminary version of this paper appears in *Proc. National Conference on Artificial Intelligence (AAAI '96), pp. 1313–1319.*

## Appendix A. Proof of Lemma 3.6

Recall that all that remains in the proof of Lemma 3.6 is to deal with the case that $\Pr(V_3|V_2 \cap V_1) = \Pr(U_3|U_2 \cap U_1)$ and $\Pr(V_2|V_1) = \Pr(U_2|U_1)$, and hence $\Pr(V_3 \cap V_2|V_1) = \Pr(U_3 \cap U_2|U_1)$.

Before proceeding with the proof, it is useful to collect some general facts about Pr. A set $U$ is said to be *standard* if $U$ is a subset of one of $\{w_1, w_2, w_3\}$, $\{w_4, w_5, w_6\}$, $\{w_7, w_8, w_9\}$, or $\{w_{10}, w_{11}, w_{12}\}$. A real number $a$ is said to be *relevant* if there exists some standard $U$ and some arbitrary $V$ such that $a = \Pr(V|U)$. Notice that even if $U \neq \emptyset$ is nonstandard, then, taking $U'$ to be the standard subset of $U$ which has the greatest weight, then $|\Pr(V|U) - \Pr(V|U')| < .002$. (This is the reason that the weights are multiplied by factors such as $10^4$, $10^8$, and $10^{18}$.) Thus, for any subsets $V$ and $U$ of $W$, we have that $\Pr(V|U)$ is close to a relevant number (where "close" means "within .002").

Call a triple $(U, V, V')$ of subsets of $W$ *good* if $\mathrm{Bel}_0(V' \cap V|U) = \mathrm{Bel}_0(V'|V \cap U) \times \mathrm{Bel}_0(V|U)$. Clearly if both $(U_1, U_2, U_3)$ and $(V_1, V_2, V_3)$ are good, then the lemma holds. Notice that if $(U, V, V')$ is not good, then $U \supseteq \{w_{10}, w_{11}, w_{12}\}$ and $f(V \cap \{w_{10}, w_{11}, w_{12}\}) \neq f'(V \cap \{w_{10}, w_{11}, w_{11}\})$, which means that $V \cap \{w_{10}, w_{11}, w_{12}\}$ must contain one of $w_{10}$ and $w_{11}$, but not both, and thus must be one of $\{w_{10}\}$, $\{w_{11}\}$, $\{w_{10}, w_{12}\}$, or $\{w_{11}, w_{12}\}$.

Thus, we may as well assume that at least one of $(U_1, U_2, U_3)$ or $(V_1, V_2, V_3)$ is not good. In that case, I claim that one of the following must hold:

- $\mathrm{Bel}_0(V_3 \cap V_2|V_1) = Bel(V_3|V_2 \cap V_1) = \mathrm{Bel}_0(U_3|U_2 \cap U_1) = \mathrm{Bel}_0(U_3 \cap U_2|U_1) = 0$

- $U_3 \cap U_2 \cap U_1 = U_2 \cap U_1$ and $V_3 \cap V_2 \cap V_1 = V_2 \cap V_1$

- $f(U_1) = f(V_1)$ and $f(U_1 \cap U_2) = f(V_1 \cap V_2)$

In the first case, we have already seen that the lemma holds. In the second case, we have $\mathrm{Bel}_0(V_3 \cap V_2|V_1) = \mathrm{Bel}_0(V_2|V_1)$, $\mathrm{Bel}_0(U_3 \cap U_2|U_1) = \mathrm{Bel}_0(U_2|U_1)$, and $\mathrm{Bel}_0(V_3|V_2 \cap V_1) = \mathrm{Bel}_0(U_3|U_2 \cap U_1) = 1$, so the lemma is easily seen to hold. Finally, in the third case, notice that since $\Pr(U_2 \cap U_3|U_1) = \Pr(V_2 \cap V_3|V_1)$, we must also have that $f(U_1 \cap U_2 \cap U_3) =$

$f(V_1 \cap V_2 \cap V_3)$. Moreover, it is easy to see that all these equalities must hold if $f$ is replaced by $f'$. Again, the lemma immediately follows.

To prove the claim, for definiteness, assume that $(U_1, U_2, U_3)$ is not good (an identical argument works if $(V_1, V_2, V_3)$ is not good). From the characterization above of triples that are not good, it follows that $f(U_1 \cap U_2) = a \times 10^{18} + b$ and $f(U_1) = 19 \times 10^{18} + c$, where $a \in \{2, 3, 16, 17\}$ (depending on $U_2 \cap \{w_{10}, w_{11}, w_{12}\}$), and both $b, c < 20 \times 10^8$. Clearly, the relevant number closest to $\Pr(U_2|U_1)$ is $a/19$. Since $\Pr(V_2|V_1) = \Pr(U_2|U_1)$ by assumption, $\Pr(V_2|V_1)$ is also close to $a/19$. Thus, we must have that $f(V_1 \cap V_2) = a \times 10^k + b'$ and $f(V_1) = 19 \times 10^k + c'$, where $k \in \{0, 4, 8, 18\}$. In fact, it is easy to see that $k$ is either 8 or 18, since there are no relevant numbers of the form $a/19$ (for $a \in \{2, 3, 16, 17\}$) that are close to $\Pr(V|U)$ if $U \subseteq \{w_1, w_2, w_3, w_4, w_5, w_6\}$. In addition, if $k = 18$, then $b', c' < 20 \times 10^8$, while if $k = 8$, then $b', c' < 20 \times 10^4$. By standard arithmetic manipulation, we have that

$$10^{18}(ac' - 19b') + 10^k(19b - ac) + (bc' - b'c) = 0.$$

If $k = 8$, then it is easy to see that we must have

$$ac' - 19b' = 0, \ 19b - ac = 0 \text{ and } bc' - b'c = 0, \tag{11}$$

while if $k = 18$, then we must have

$$19(b - b') + a(c' - c) = 0 \text{ and } bc' - b'c = 0. \tag{12}$$

Now comes a case analysis. First suppose that $k = 8$. Then we must have $b' = c' = 0$, since if $c' \neq 0$, then from (11) we have that $b'/c' = a/19$, and it is easy to see that there do not exist sets $T_1$ and $T_2$ such that $f(T_1) = b'$, $f(T_2) = c'$, and $b'/c' = a/19$, with $b', c' \leq 20 \times 10^4$. Thus, it follows that $\Pr(U_2|U_1) = \Pr(V_2|V_1) = a/19$. Moreover, we must have $V_1 = \{w_7, w_8, w_9\}$ and $V_2 \cap V_1$ either $\{w_7\}$ or $\{w_8, w_9\}$, depending on $a$. It follows that $\Pr(V_3|V_2 \cap V_1)$ must be one of $\{0, 1/2, 1\}$. Since $\Pr(U_3|U_2 \cap U_1) = \Pr(V_3|V_2 \cap V_1)$, we must have that $\Pr(U_3|U_2 \cap U_1) \in \{0, 1/2, 1\}$. Since $U_2 \cap U_1$ contains exactly one of $w_{10}$ and $w_{11}$, it is easy to see that $\Pr(U_3|U_2 \cap U_1)$ cannot be $1/2$. If $\Pr(U_3|U_2 \cap U_1) = \Pr(V_3|V_2 \cap V_1) = 0$, then $U_3 \cap U_2 \cap U_1 = V_3 \cap V_2 \cap V_1 = \emptyset$, and we must have $\mathrm{Bel}_0(U_3 \cap U_2|U_1) = \mathrm{Bel}_0(V_3 \cap V_2|V_1) = 0$, so the claim follows. On the other hand, if $\Pr(U_3|U_2 \cap U_1) = \Pr(V_3|V_2 \cap V_1) = 1$, then $U_3 \cap U_2 \cap U_1 = U_2 \cap U_1$ and $V_3 \cap V_2 \cap V_1 = V_2 \cap V_1$, and the claim again follows.

Now suppose $k = 18$. If $c = c'$, then by (12), we must have that $b = b'$. It immediately follows that $f(U_1) = f(V_1)$ and $f(U_1 \cap U_2) = f(V_1 \cap V_2)$, so the claim holds. Thus, we can suppose $c \neq c'$. Suppose that $c \neq 0$ (an identical argument works if $c \neq 0$). Then there exists some $x \neq 1$ such that $c = xc'$. Since $bc' - b'c = 0$, it follows that $b = xb'$. Substituting $xb'$ for $b$ and $xc'$ for $c$ in (12), we get that $(1 - x)b'/(1 - x)c' = a/19$, from which it follows that $b'/c' = a/19$. Moreover, we also get that either $b = c = 0$ or $b/c = a/19$. It is easy to check that $a$ must be either 3 or 16. If $b/c = a/19$, then we must have $b = b'$ and $c = c'$. As we have seen, this suffices to prove the claim. Thus, we can assume that $b = c = 0$. But this means that $U_1 = \{w_{10}, w_{11}, w_{12}\}$, and that $U_1 \cap U_2$ is either $\{w_{10}\}$ or $\{w_{11}, w_{12}\}$. It follows that the only possibilities for $\Pr(U_3|U_2 \cap U_1)$ are 0, 1/8, 7/8, or 1. It is easy to see that $\Pr(V_3|V_2 \cap V_1)$ cannot be 1/8 or 7/8, while the cases where it is either 0 or 1 are easily taken care of, as above.

This completes the proof of the claim and of the lemma. $\square$